\documentclass[letterpaper, 10 pt, conference]{ieeeconf}  

\IEEEoverridecommandlockouts                              

\usepackage{graphicx} 
\usepackage{xcolor}
\usepackage{bm} 
\usepackage{amssymb}  
\let\labelindent\relax
\usepackage{enumitem}
\usepackage{subfigure}

\usepackage{cite}
\usepackage{amsmath,amssymb,amsfonts}
\usepackage{algorithmic}
\usepackage{graphicx}
\usepackage{textcomp}
\usepackage{xcolor}
\usepackage{mathtools}
\usepackage{amsmath}
\usepackage{amsfonts}
\usepackage{tabularx}
\usepackage{multirow}
\usepackage{multicol}
\usepackage{float}
\usepackage{tabularx}
\usepackage{threeparttable}
\usepackage{booktabs}
\usepackage{longtable}
\usepackage{booktabs}
\usepackage{makecell}
\usepackage{verbatim}
\usepackage{wrapfig}
\usepackage{xspace}

\usepackage{hyperref}
\hypersetup{
    colorlinks=true,
    linkcolor=blue, 
    filecolor=magenta,  
    urlcolor=cyan,
    pdftitle={Overleaf Example},
    pdfpagemode=FullScreen,
    }

\title{\LARGE \bf
Sim2Real$^\textbf{2}$: Actively Building Explicit Physics Model for Precise Articulated Object Manipulation
}
\author{Liqian Ma$^{\dag1}$, Jiaojiao Meng$^{\dag2}$, Shuntao Liu$^{\dag3}$, Weihang Chen$^{1}$, Jing Xu*$^{1}$, and Rui Chen*$^{1}$
\thanks{$^{1}$Liqian Ma, Weihang Chen, Jing Xu, and Rui Chen are with the State Key Laboratory of Tribology, the Beijing Key Laboratory of Precision/Ultra-Precision Manufacturing Equipment Control, and the Department of Mechanical Engineering, Tsinghua University, Beijing 100084, China. Corresponding authors: Jing Xu {\tt\small jingxu@tsinghua.edu.cn}, Rui Chen {\tt\small chenruithu@mail.tsinghua.edu.cn}.}%
\thanks{$^{2}$Jiaojiao Meng is with the School of Modern Post, Beijing University of Posts and Telecommunications, Beijing 100876, China.}%
\thanks{$^{3}$Shuntao Liu is with AVIC Chengdu Aircraft Industrial (Group) Co., Ltd, Chengdu, Sichuan 610092, China.}
\thanks{$^\dag$equal contribution}
}

\begin{document}

\maketitle
\thispagestyle{empty}
\pagestyle{empty}

\begin{abstract}

Accurately manipulating articulated objects is a challenging yet important task for real robot applications. In this paper, we present a novel framework called {Sim2Real$^\textbf{2}$} to enable the robot to manipulate an unseen articulated object to the desired state precisely in the real world with no human demonstrations. We leverage recent advances in physics simulation and learning-based perception to build the interactive explicit physics model of the object and use it to plan a long-horizon manipulation trajectory to accomplish the task. However, the interactive model cannot be correctly estimated from a static observation. Therefore, we learn to predict the object affordance from a single-frame point cloud, control the robot to actively interact with the object with a one-step action, and capture another point cloud.
Further, the physics model is constructed from the two point clouds. Experimental results show that our framework achieves about $70\%$ manipulations with $<30\%$ relative error for common articulated objects, and $30\%$ manipulations for difficult objects. Our proposed framework also enables advanced manipulation strategies, such as manipulating with different tools.
Code and videos are available on our project webpage \href{https://ttimelord.github.io/Sim2Real2-site/}{https://ttimelord.github.io/Sim2Real2-site/}

\end{abstract}

\section{INTRODUCTION}

The ability to manipulate articulated objects such as opening cabinets and turning faucets is critical for household robot use cases. Because manipulation is a sequence of different actions corresponding to different object states, it is difficult for neural networks to learn the correlation, even with hundreds of successful demonstrations and millions of interactions~\cite{mu2021maniskill}. For human beings, the manipulation does not only involve the responding action to perception, which is the case for visual policy networks, but also involves motor imagery and mental simulation~\cite{johnson2000thinking}. And the development of the mental model and cognition is closely dependent on actions~\cite{von2007action, hofsten2009action}. By actively interacting with the environment and changing the state, intelligent agents can gather additional information and model the world more accurately.

\begin{figure}[thpb]
      \centering
      \includegraphics[width=7.8cm]{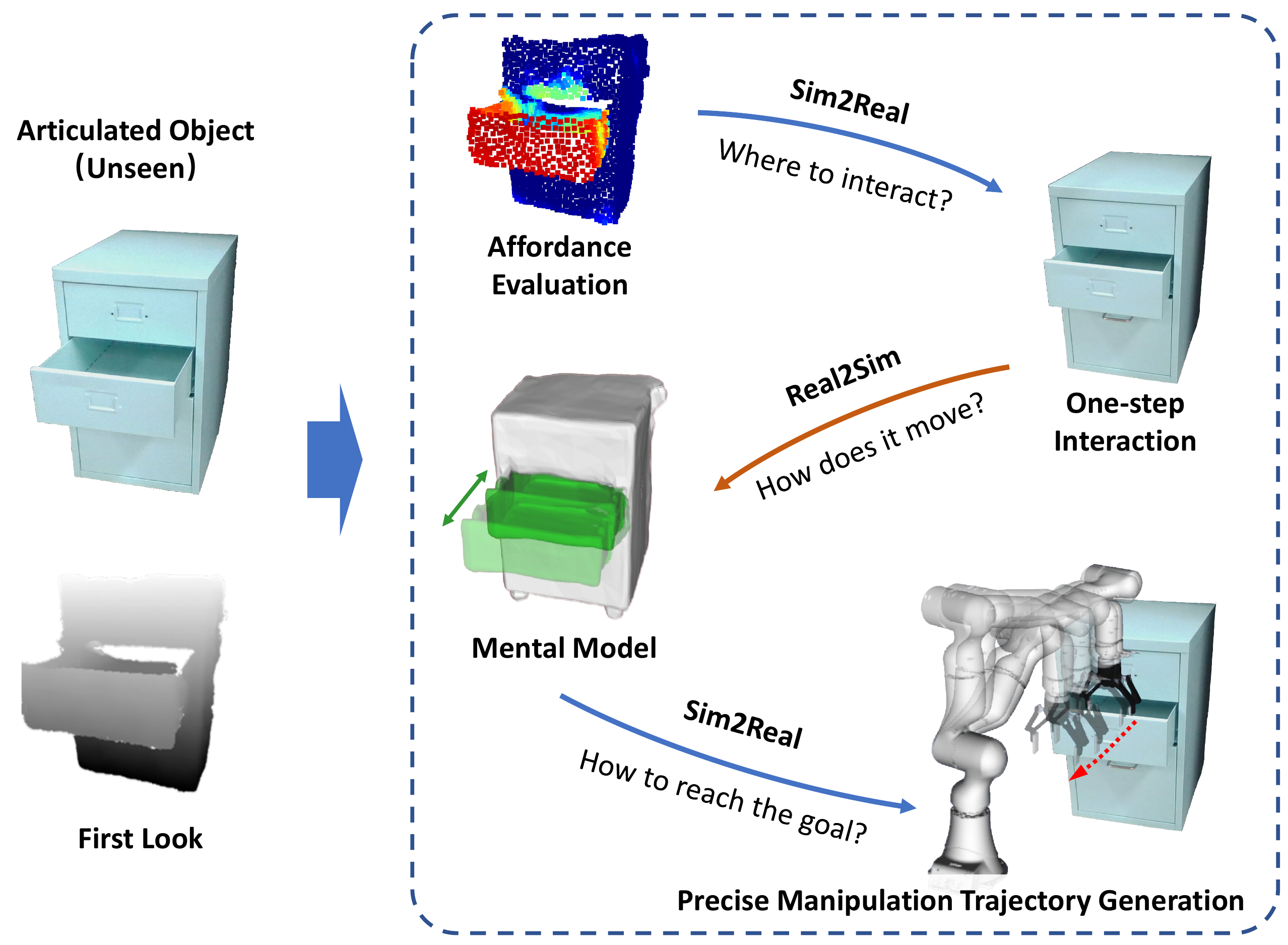}
      \caption{\textbf{Sim2Real$^\textbf{2}$} is a robot learning framework for precise articulated object manipulation in the real world. It builds the mental model of the unseen target object through one-step active interaction and uses the model to generate a long-horizon manipulation trajectory.}
      \label{fig:teaser}
      \vspace{-5mm}
\end{figure}

In this paper, we propose a robot learning framework for precise articulated object manipulation in the real world called \textbf{Sim2Real$^\textbf{2}$} (Fig.~\ref{fig:teaser}), where we use a physics simulator as the \textit{mental model} of robots. The framework first learns an action affordance estimation network in the \textbf{simulation}, which takes a partial point cloud as input and predicts a one-step action to change the object state. Because affordance estimation is only attributed to the object and not related to the robot, it is easier to learn and has better generalizability to novel objects. Secondly, we feed the observed point cloud from the real depth sensor to the learned network, execute the predicted action on the \textbf{real} robot and capture another point cloud after execution. Thirdly, we use a learning-based method to construct the explicit physics model of the single object instance in the \textbf{simulation}. Given the goal state of the object, we generate a long-horizon manipulation trajectory using model predictive control (MPC). Finally, we execute the trajectory on the \textbf{real} robot to manipulate the object to the goal state.
\begin{figure*}[thpb]
    \centering
    \includegraphics[width=17cm]{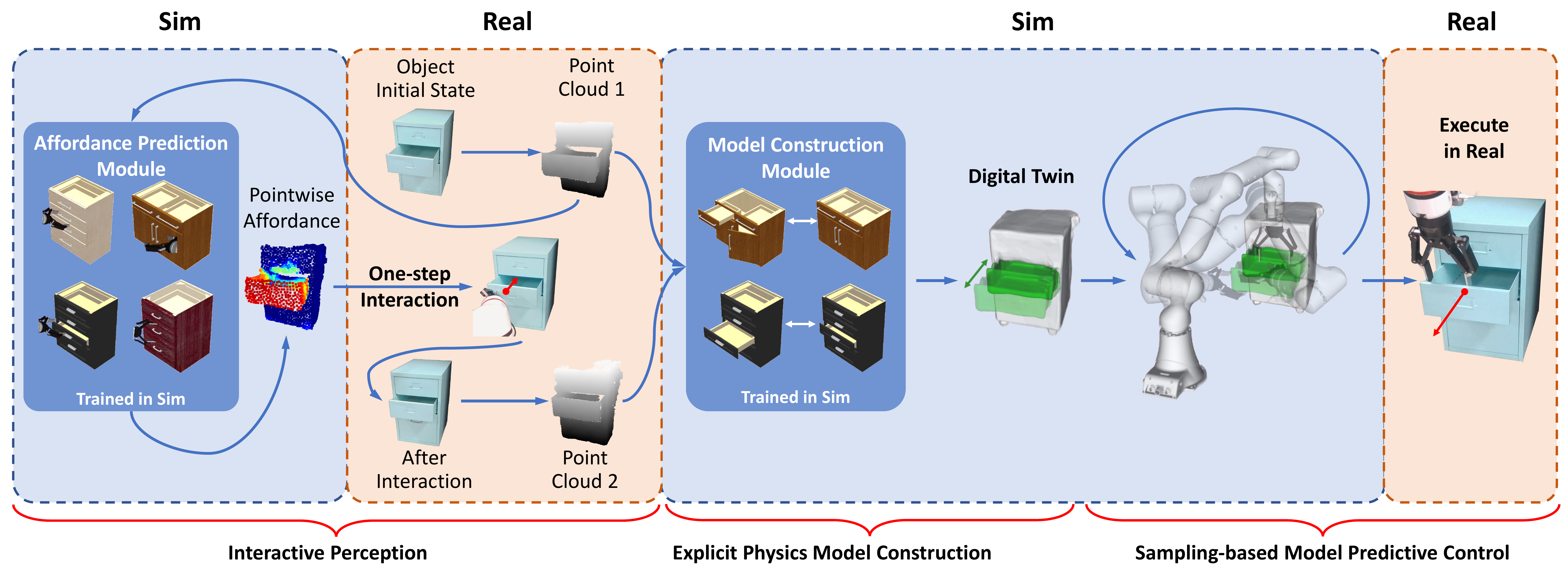}
    \caption{{Sim2Real$^\textbf{2}$} framework overview. Our framework consists of three phases: Given a partial point cloud of an unseen articulated object, in the Interactive Perception phase, we train an affordance prediction module in simulation and use its prediction to change the object's joint state through one-step interaction; in Explicit Physics Model Construction phase, we build an interactive model from the two point clouds; in Sampling-based Model Predictive Control phase, we use the model to plan a long-horizon trajectory in simulation and finally execute the trajectory on the real robot to complete the task.}
    \label{fig:method}
    \vspace{-5mm}
\end{figure*}
Compared with the latent world model based on neural networks~\cite{hafner2019dream, hafner2020mastering, wu2022daydreamer}, we only require one interaction in the real world, and the model is guaranteed to generalize and extrapolate to unseen actions. We achieve this by introducing the structured physics prior of articulated objects to the model-building procedure.

The key contributions of the paper are as follows:

\begin{enumerate}[label=(\arabic*)]
    \item we propose a robot learning framework for precise articulated object manipulation.
    \item we quantitatively evaluate the effectiveness of the proposed framework through real experiments on 9 articulated objects of 3 different categories.
    \item we show that our framework can support advanced manipulation skills, such as manipulating with tools.
\end{enumerate}

\section{RELATED WORK}

\noindent{\textbf{Building Transition Model}}
Model-based reinforcement learning(MBRL) builds a transition model for the environment and reduces the need of environment interactions substantially~\cite{polydoros2017survey, wang2019benchmarking,moerland2020model}. Given the prevalence of deep learning, high-capacity neural networks are usually used as the transition model~\cite{hafner2019dream, hafner2020mastering, wu2022daydreamer}. But because the networks contain minimal prior of the environment, still a large number of training samples are required to improve the generalizability~\cite{plaat2020model}. In our work, we focus on articulated object manipulation, so we introduce the prior through an explicit physics model and decrease the  number of required samples to 1.
Moreover, the generalizability of the explicit physic model guarantees that while we only use a simple action to collect the sample, the built model can be used for long-horizon complex trajectory planning composed of unseen actions.

\noindent{\textbf{Sim2Real for Robot Learning}}
Physics simulation plays a key role in robot learning as it allows large-scale parallelism, reduces the training cost, and avoids potential damage to robots and researchers~\cite{james2020rlbench, zhu2020robosuite, makoviychuk2021isaac}. Most existing methods train an RL policy in the simulation and transfer the policy to the real robot~\cite{sadeghi2018sim2real,hofer2021sim2real, dimitropoulos2022brief}. However, it entangles two difficult problems: the policy's generalizability to novel objects in the real world, and the domain gap between simulation and the real world. In our work, we disentangle the two problems by using simulation in two different ways: firstly, we learn a simple policy on large-scale object assets to interact with the object and build the model. The simplicity of the policy leads to better generalizability. Secondly, we use MPC to generate the manipulation trajectory on the built model of the single object instance. Because the diversity of objects is avoided, it reduces the number of interactions substantially and improves the manipulation accuracy.

\noindent{\textbf{Articulated Object Manipulation}}
Manipulation of articulated objects is important to real robot applications and remains an open research problem due to the objects' complex kinematic constraints. Learning-based methods have shown promising progress on this problem, but most of them only evaluate the effectiveness in the simulation~\cite{gadre2021act, mu2021maniskill}. Eisner~\textit{et al.} proposed to manipulate the objects based on the per-point articulation flow estimation, but it is only applicable to suction grippers and does not infer the structure of the object. Our work is most related to SAGCI-system~\cite{lv2022sagci}, which uses a learning-augmented differentiable simulation to produce manipulation policies. Our work differs from theirs in three key ways:
(1) our method does not require differentiable physics simulation so that it can be adaptable to all existing physics simulators and does not require persistent contact between the robot and the object; (2) we carried out an extensive quantitative evaluation on real robot experiments; (3) our framework enables advanced manipulation policies, such as manipulating with tools.

\section{METHOD}

\subsection{Overview}
An overview of our framework is shown in Fig.~\ref{fig:method}.
The goal of our work is to manipulate real-world articulated objects to desired joint states. When the robot is deployed in the real world, it can use sensors like RGB-D cameras to perceive the articulated object. The observation is usually partial because of self-occlusion. 
So our learning framework takes point clouds generated from single-view depth images as input.

The information about the articulated object is not complete in a single-frame observation. For example, when humans first look at a kitchen door, it is hard to tell whether it has a rotating hinge or a sliding hinge. But after the door is moved, humans can use the information contained in the two observations to infer which kind of hinge the door has and where the hinge is. With an initial observation (which is a single-view point cloud) as input, the \textbf{Interactive Perception} module~(\ref{Interactive Perception}) proposes an action which can change the joint state of the articulated object. Then the action is executed on the object. Another frame of point cloud is acquired.

With the two observations of the articulated object, the \textbf{Explicit Physics Model Construction} module~(\ref{Explicit Physics Model Construction}) can construct a digital twin of it. We use Unified Robot Description Format (URDF) to represent the constructed model. URDF files can be easily loaded into all kinds of multi-body physics simulators, such as Sapien~\cite{xiang2020sapien}. In the simulator, the robot can interact with the articulated object, which forms a mental model of reality. 

Once the mental model is constructed, we use \textbf{Sampling-based Model Predictive Control}~(\ref{Sampling Based Model Predictive Control}) to plan how to change the state of the articulated object to a desired one. Given one specific initial state and one goal state of a specific model, we no longer require the generalizability, so we choose to utilize the iCEM method~\cite{pinneri2020sample}, a sampling-based zeroth-order trajectory optimizer in an MPC framework, to search for the goal trajectory to accomplish the task.

\subsection{Interactive Perception}
\label{Interactive Perception}
The robot gets a better understanding of the articulated object by actively interacting with it. Because the robot only has a single-frame point cloud $\bm{\mathcal{P}}_0\in\mathbb{R}^{N\times3}$ as observation at the beginning, it has to infer an action $\bm{a}^*$ to change the object's joint state $s_0$ to get more information.

In our work, we choose Where2act~\cite{mo2021where2act} for affordance predicting in the Interactive Perception module. It predicts object affordance related to the end-effector. Other works like~\cite{eisner2022flowbot3d,xu2021umpnet} can also be used in our framework.

In~\cite{mo2021where2act}, the algorithm has an Actionability Scoring Module which predicts actionability score $a_p$ over all the points; 
the Action Proposal Module proposes actions on a particular point; the Action scoring module predicts success likelihood score of the proposed actions.

In the Where2act method, only a flying gripper is considered. The primitive actions are parametrized by gripper pose in $SE(3)$ space. It does not consider the kinematic structure of the robot, which increases the difficulty of execution in the real world because motion planning may fail. 
To solve this problem, we choose $n_p$ points with the highest actionability scores as candidate points. For each point, we select $n_a$ actions with the highest success likelihood scores from the proposed actions. We use motion planning to try to generate joint trajectories for the actions one by one until a successful one is found. We empirically find that this method improves the success rate for the motion planner because the action with the highest success likelihood is often out of the dexterous workspace of the robot.
Finally, we execute the planned trajectory on the real robot, which changes the articulated object's joint state from $s_0$ to $s_1$. Then the agent can observe another point cloud $\bm{\mathcal{P}}_1$.

\subsection{Explicit Physics Model Construction}
\label{Explicit Physics Model Construction}
Building an explicit model of an articulated object is difficult because only if the geometries of all parts and kinematic relationships between connected parts are both figured out can the model of the articulated object be constructed.

Our work focuses on the Sim2Real of articulated objects, which needs to use the model trained in physical simulation to reconstruct the interactive digital twins of the articulated objects in the real world and use it in the physics simulation to interact with the robot. 
In our work, we have two assumptions for the constructed physics model: (1) the articulated object only contains a single prismatic or revolute joint; (2) the base link of the articulated object is fixed.

We choose Ditto~\cite{jiang2022ditto} to construct the physical model explicitly. Given visual observations before and after the interaction $\bm{\mathcal{P}}_0$ and $\bm{\mathcal{P}}_1$, Ditto uses structured feature grids and unified implicit neural representation to construct part-level digital twins of articulated objects. Different from the original work where a multi-view fused point cloud is used, we use a single-view point cloud as input, which is more consistent with real robot application settings. Furthermore, we simulate the depth sensor noise when generating training data to narrow the domain gap~\cite{zhang2022close}. After we train the Ditto on simulated data, we use the trained model on the real two-frame point clouds to generate the implicit neural representation and extract the meshes.
The explicit physics model is represented as URDF, which can be loaded into existing physics simulators.
We further perform convex decomposition using VHACD~\cite{mamou2009simple} before importing the meshes to the physics simulator, which is essential for realistic physics simulation of robot interaction.

\subsection{Sampling-based Model Predictive Control}
\label{Sampling Based Model Predictive Control}
Having an explicit physics model and a target joint state $s_{target}$ of the articulated object, the agent needs to search for a trajectory that can change the current joint state $s_{initial}=s_{1}$ to $s_{target}$. The expected relative joint movement is $\Delta s_{target} = s_{target} - s_{initial}$. 
In our work, we use the iCEM method~\cite{pinneri2020sample} to search for a feasible long-horizon trajectory to complete the task:

Trajectory length $T\in\mathbb{N}^+$ denotes the maximum time steps in a trajectory. 
At each time step $t (t<T)$, the action of the robot $\bm{a}_t\in \mathbb{R}^d$ is the incremental value of the joint position, where $d$ is the number of degrees of freedom (DOF) of the robot. The population $N$ denotes the number of samples sampled in each CEM-iteration. Planning horizon $h$ determines the number of time steps the robot plans in the future at each time step. The top $K$ samples according to rewards compose elite-set, which is used to fit means and variances of a new Gaussian distribution. Please refer to~\cite{pinneri2020sample} for details of the algorithm.

To speed up the searching process, we use dense rewards to guide the trajectory optimization.
The reward function consists of the following terms:

\noindent (1) \emph{success reward}
\begin{equation*}
            r_{success}=\begin{cases}\omega_s, \quad & \text{if} \left|s_{target}-s_t\right|<\epsilon\\
                                     0, \quad & \text{else}\end{cases}
        \end{equation*}
\noindent where $s_t$ denotes the joint state at current time step $t$, and $\epsilon$ is a predefined threshold.

\noindent (2) \emph{approaching reward}
\begin{equation*}
            r_{target} = -\omega_t*(s_{target}-s_{t})/(s_{target}-s_{initial}) 
        \end{equation*}
\noindent         This reward encourages $s_{t}$ to converge to $s_{target}$.

\noindent (3) \emph{contact reward}
\begin{equation*}
            r_{contact} = 
            \begin{cases}
                    \omega_{contact}, & \text{if}\quad \dfrac{\left| s_{t}-s_{target} \right|}{\left| s_{target}-s_{initial} \right|}<1\\
                    -\omega_{collision}, & \text{if unexpected collision happens} \\
                    0, & \text{else}
            \end{cases}
        \end{equation*}
\noindent         This reward encourages the robot to have first contact with the object in the correct direction and to keep in contact with the object when moving the part. Also, this reward tries to prevent parts other than the fingertip or the target part of the object from colliding.
        
\noindent (4) \emph{distance reward}
\begin{equation*}
            r_{dist} = \omega_d * || \bm{p}_{part} - \bm{p}_{grasp} ||^2
        \end{equation*}
\noindent         This reward encourages the gripper to get closer to the target part of the object, where $\bm{p}_{part}, \bm{p}_{grasp} \in \mathbb{R}^3$ denotes the position of the geometry center of the target part and the grasp center of the gripper in Cartesian space, respectively.
        
\noindent (5) \emph{regularization reward}
 \begin{equation*}
            r_{reg} = -\sum_{i=0}^d{(\omega_a * a_i + \omega_v * v_i)}
        \end{equation*}
 \noindent        This reward is a regularization reward that discourages the robot to move too fast or move to an unreasonable configuration. $a_i$ and $v_i$ denote the acceleration and velocity of the $i$th joint respectively.

Once the manipulation trajectory is generated, we execute the trajectory on the real robot.

\section{EXPERIMENTS}

We examine the effectiveness of the proposed method to manipulate the articulated objects precisely. We first perform a large amount of real articulated object manipulation experiments to quantitatively evaluate the accuracy of our method. Then, ablation studies are conducted to validate the effectiveness of the parts of our method. Finally, we evaluate the flexibility of our method by completing articulated manipulation tasks with tools.

We choose 3 categories of articulated objects for experiments, which are drawers, faucets, and laptops. 
We use Sapien physics simulator~\cite{xiang2020sapien} to collect data for training of the Interactive Perception module and Explicit Physics Model Construction module and create simulation environments for Sampling-based Model Predictive Control.

\begin{figure}[t]
\centering
 \subfigure[]
 {
  \includegraphics[height=3.6cm]{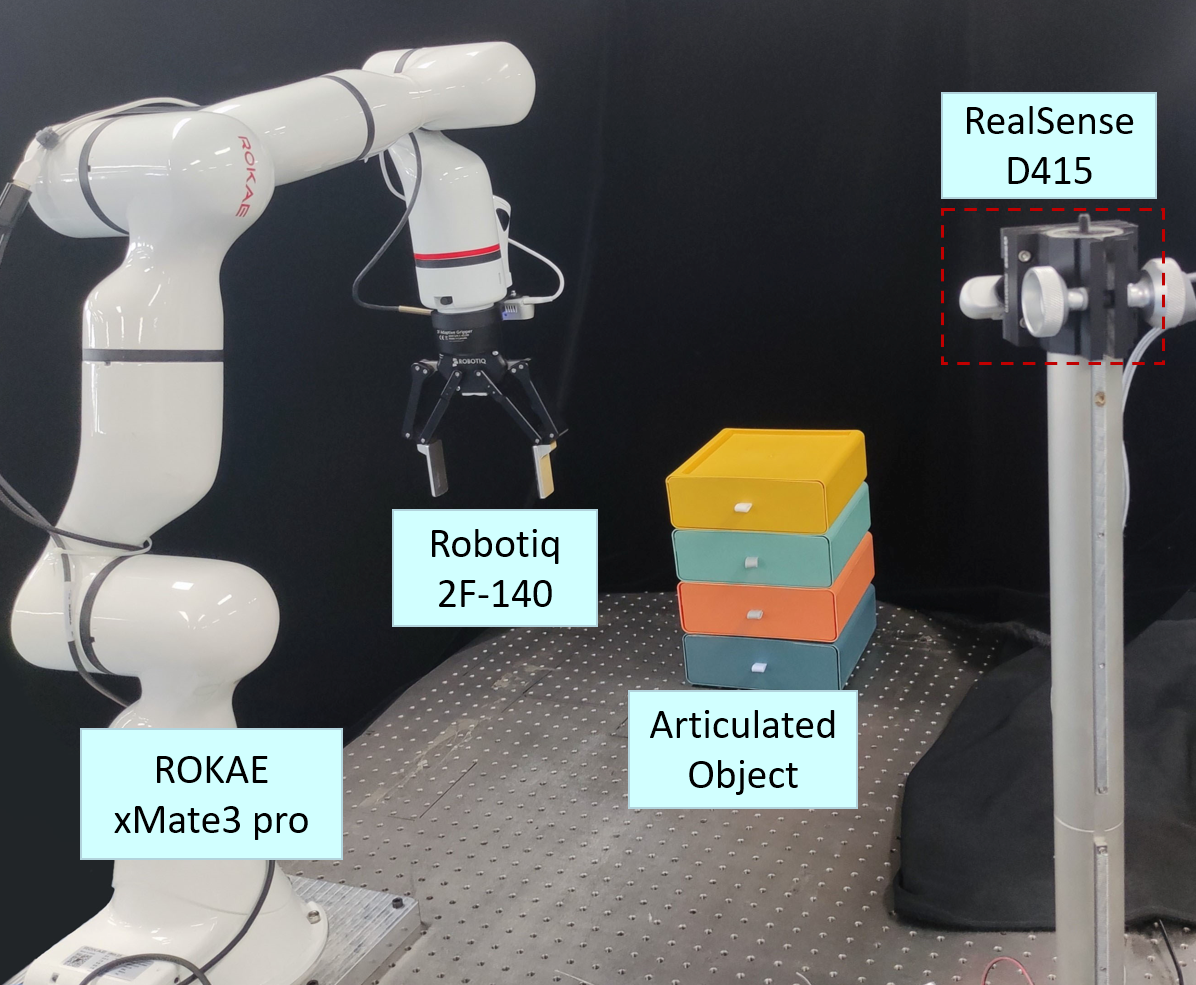}
 \label{fig:setup}
 }
 \hspace{-4mm}
  \subfigure[]
 {
  \includegraphics[height=3.6cm]{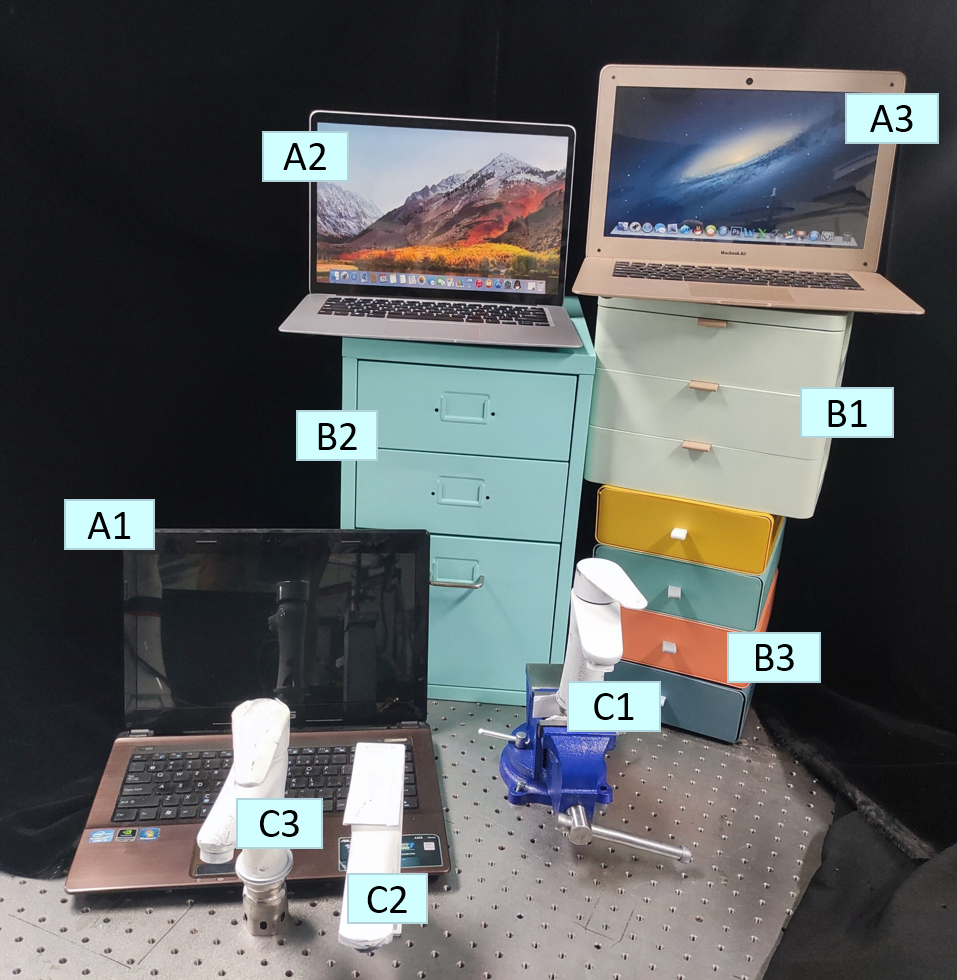}
 \label{fig:objects}
 }
 \caption{Real-world experimental setup (a) and the manipulated objects (b). There are 3 categories of articulated objects for our real-world experiment.}
 \vspace{-5mm}
\end{figure}

\subsection{Data Collection and Training}

\noindent{}\textbf{Interactive Perception.}  We choose 35 drawers, 8 faucets, and 7 laptops from the Partnet Mobility dataset~\cite{xiang2020sapien} for data collection. We use pushing primitive when collecting data on drawers and laptops, and pushing-left primitive when collecting data on faucets. The 3 categories are trained separately for the difference in action primitives. 70000, 96000, and 14000 samples are collected for drawers, faucets, and laptops, respectively. The way of data collecting is similar to~\cite{mo2021where2act}. Here are some differences: (1) we randomly set the target joint state (2) The azimuth of the camera is randomly sampled from $[-60^\circ, +60^\circ)$ and the altitude is randomly sampled from $[15^\circ, 45^\circ)$. (3) we move the center of the point cloud to the origin instead of translating the point cloud by a certain value, which is beneficial for deployment in real environment because objects may appear at different distances from the camera. We downsample the object point cloud to 2000 points before using it as the network input. 

\noindent{}\textbf{Explicit Physics Model Construction.} For drawers and faucets, we choose 14 and 8 objects from Shape2Motion datasets~\cite{wang2019shape2motion}. For laptops, we choose 5 objects from PartNet-Mobility dataset~\cite{xiang2020sapien} because the joint limits of laptops in PartNet-Mobility dataset are more reasonable than in Shape2Motion datasets. The original meshes are not water-tight in the PartNet-Mobility dataset, which cannot be used to compute point occupancy, so we use ManifoldPlus~\cite{huang2020manifoldplus} to fix the meshes. We reimplement the data collection code using Sapien simulator to keep consistency. 
When collecting data, the object is fixed on the origin and then randomly rotated around the z-axis by $[-60^\circ, +60^\circ)$. The camera is on a sphere centered on the object's center. Azimuth and Altitude of the camera are randomly sampled from $[-60^\circ, +60^\circ)$ and $[15^\circ, 45^\circ)$. 10000 samples are collected for each category. We downsample the object point clouds to 8192 points. The 3 categories are trained jointly.

\subsection{Real World Articulated Object Manipulation}
Fig.~\ref{fig:setup} shows the setup of our real experiments, which consists of an optical table, a 7-DOF robot (ROKAE xMate3Pro) with a 2-finger gripper (Robotiq 2F-140) mounted on its end link, an RGBD camera (Intel RealSense D415) and the articulated object to be manipulated. 

We choose 9 objects in 3 categories for real object manipulation experiments (Fig.~\ref{fig:objects}). The articulated object is randomly located on the table with the base link fixed, and $s_{0}$ is randomly set. We randomly select $\Delta s_{target}$ which does not exceed the joint limit and covers both directions of possible movement.
We crop the object point cloud out of the whole scene with a bounding box. 

For parameters of the Interactive Perception module, we choose $n_p=10$ and $n_a=10$.
For parameters of the Sampling-based Model Predictive Control module, we find that $T=50$, $N=300$, $h=10$ and $K=20$ are enough to complete all the tasks. The range of incremental value of joint position is set to $[-0.05, 0.05]$. The parameters in the reward function are determined manually according to experience in the simulation environment. We set $\omega_s=20$, $\epsilon=0.005$(m or rad), $\omega_t=50$, $\omega_{contact}=10$, $\omega_{collision}=60$, $\omega_d=10$, $\omega_a=0.01$ and $\omega_v=0.03$. We use 20 processes for sampling in simulation on a computer that has an Intel Core i7-12700 CPU and an NVIDIA 3080Ti GPU. It takes 4 minutes to find a feasible trajectory.

\begin{figure}[t]
      \centering
      \includegraphics[width=7.8cm]{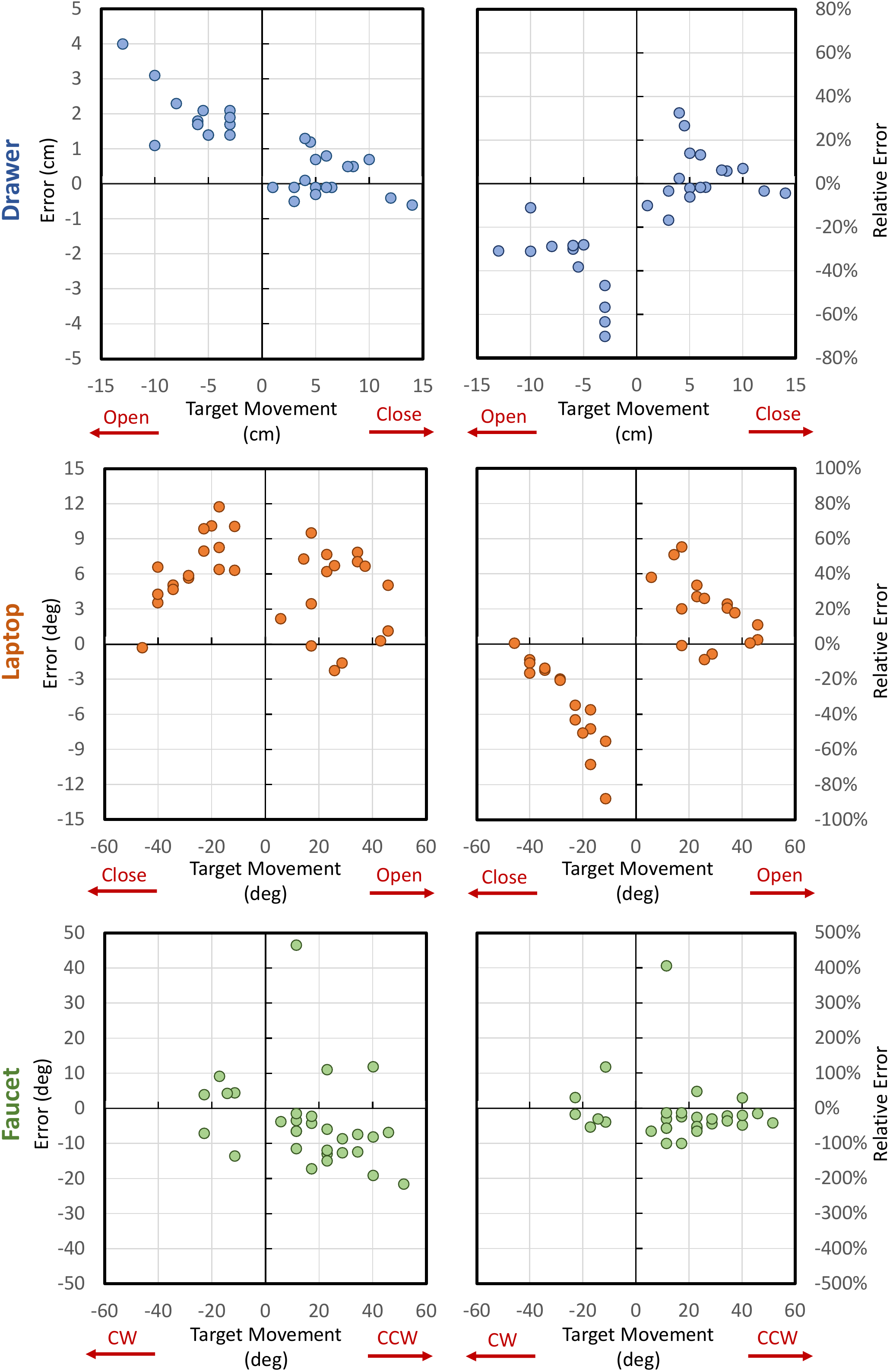}
      \caption{Results of real-world articulated object manipulation experiments. Each row shows the result of 3 categories respectively. The left column shows the error $\delta$ of the manipulation. The right column shows the relative error $\delta_r$ of the manipulation. The sign of the target movement denotes the direction of the movement (e.g. opening or closing, clockwise (CW) or counterclockwise (CCW).}
      \label{fig:real_experiement_results}
      \vspace{-2mm}
\end{figure}

\begin{figure}[thpb]
      \centering

    \subfigure[]
    {
    \centering
      \includegraphics[width=8cm]{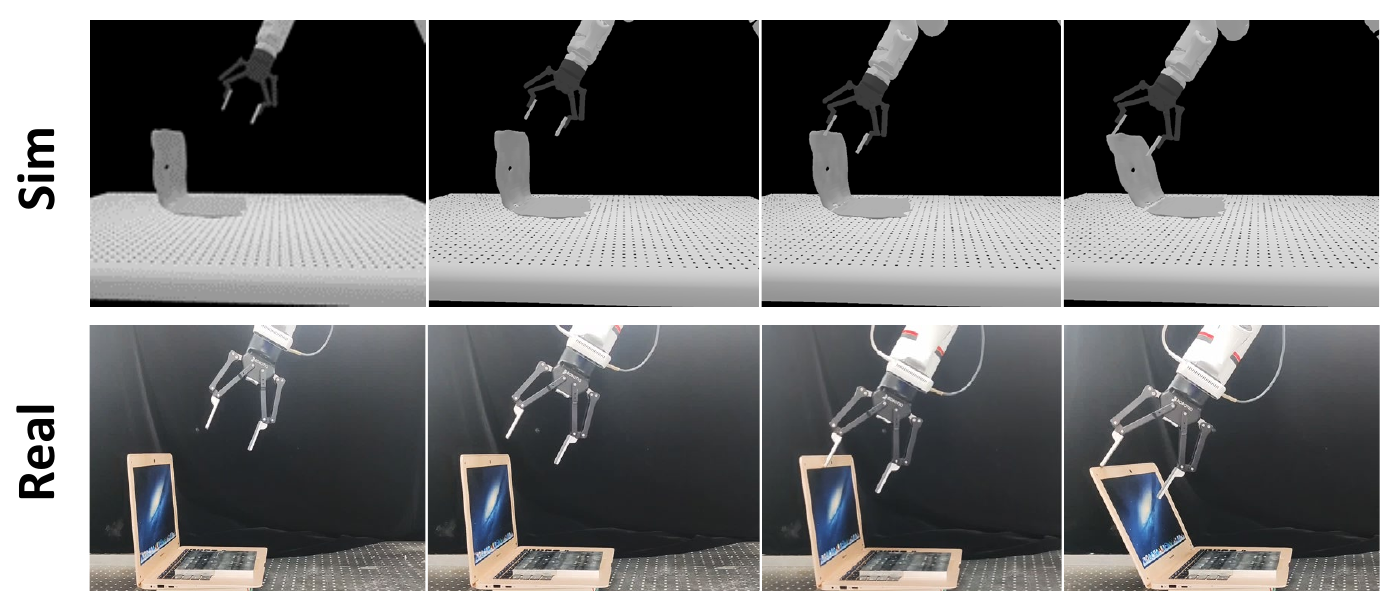}
        \label{fig:open_laptop}
     }
     \vspace{-2mm}
    \subfigure[]
    {
    \centering
      \includegraphics[width=8cm]{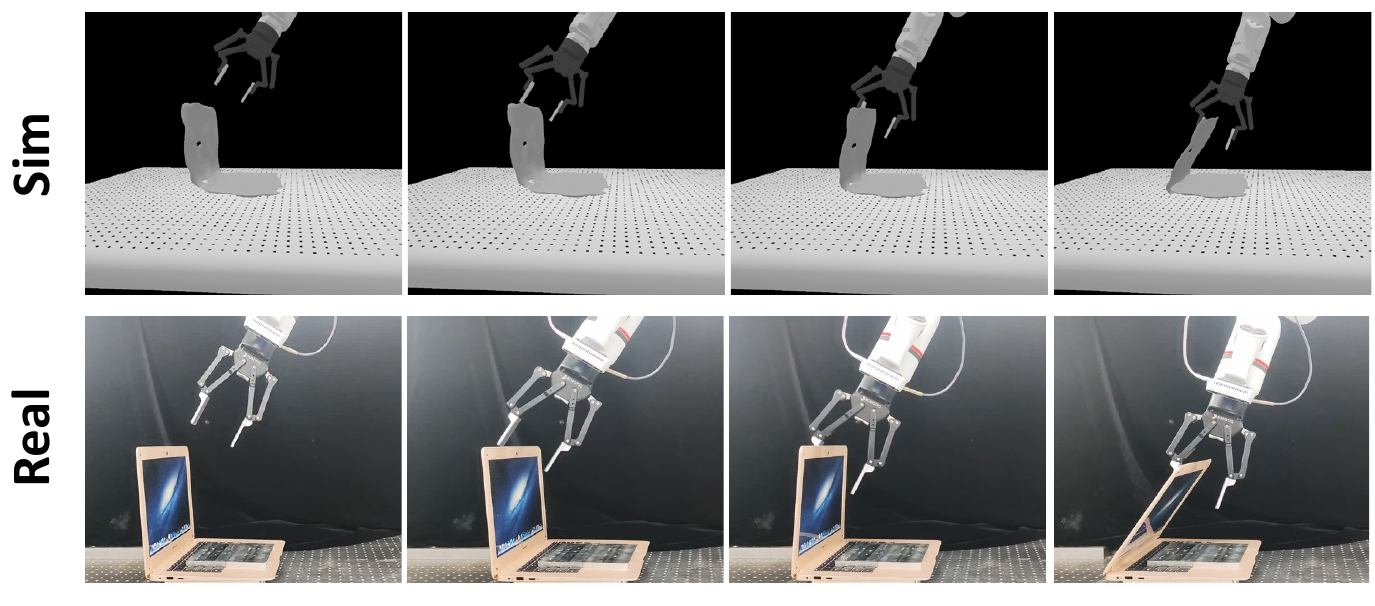}
        \label{fig:close_laptop}
      }
      \caption{Trajectories of laptop manipulation: (a) open; (b) close. The constructed digital twin precisely captures the kinematic property of the articulated object, leading to the accurate alignment of Sim and Real.}
      \label{fig:trajectory_sequences}
      \vspace{-2mm}
\end{figure}

We conduct about 30 experiments for each category. After the trajectory is executed in the real world, we measure the real joint movement $\Delta s_{real}=s_{real}-s_{initial}$ and compare it with the target joint movement $\Delta s_{target}=s_{target}-s_{initial}$. We compute the error $\delta=\Delta s_{real}-\Delta s_{target}$ and the relative error $\delta_r={\delta}/{\Delta s_{target}}\times 100\%$, results of all the experiments can be found in Fig.~\ref{fig:real_experiement_results}, and statistical results can be found in Table~\ref{tab:experimental data analysis}. Trajectories of both opening and closing the laptop are shown in Fig.~\ref{fig:trajectory_sequences}.

Among all 3 categories, the drawer has the lowest $\left|\delta_r \right|$ and the faucet has the highest $\left|\delta_r \right|$ according to Table~\ref{tab:experimental data analysis}. It is reasonable because the size of the faucet is relatively small, a minor inaccuracy in model construction or trajectory execution will result in a big error in the joint state. About $70\%$ manipulations achieve a $\left|\delta_r \right| <30\%$ for drawers and laptops, which shows the accuracy of our method.

\begin{table}[thbp]
    \centering
    \caption{Accuracy of real articulated objects manipulation}
    \resizebox{\columnwidth}{!}{%
    \begin{tabular}{cc|ccc}
    \toprule
    \multicolumn{2}{c|}{Category} &
      {Drawer} &
      {Laptop} &
      {Faucet} \\ 
     \hline
    \multicolumn{2}{c|}{Number of manipulations} & 31      & 32      & 30       \\ \hline
    \multicolumn{1}{c|}{} & \textless 10\% & 12      & 7       & 0        \\
    \multicolumn{1}{c|}{} & \textless 30\% & 22      & 20      & 9        \\
    \multicolumn{1}{c|}{\multirow{-3}{*}{\begin{tabular}[c]{@{}c@{}}Number of manipulations \\ s.t. $\left|\delta_r \right|$ \end{tabular}}} &
      \textless 50\% &
      28 &
      26 &
      19 \\ \hline
    \multicolumn{2}{c|}{Avg $\left|\delta \right|$}        & 1.15cm  & 5.69$^{\circ}$ & 10.37$^{\circ}$ \\
    \multicolumn{2}{c|}{Avg $\left|\delta_r \right|$}    & 21.81\% & 27.26\% & 56.21\%  \\ \bottomrule
    \end{tabular}%
    }
    \label{tab:experimental data analysis}
\end{table}

Errors may be caused by the following factors: 
\begin{enumerate}[label=(\arabic*)]
    \item The constructed mesh is not accurate enough, especially for the parts that are occluded. For example, the inside face of the drawer front cannot be observed by the RGBD camera, so when the digital twin is constructed, the drawer front is thicker than the real one. It causes the results of opening tasks of drawers (which has average $\left|\delta\right|$ of $2\mathrm{cm}$) to be worse than closing tasks (which have average $\left|\delta\right|$ of $0.5\mathrm{cm}$). It is worth noting that there is a relative error of over $400\%$ in turning faucet tasks. This happens because the robot touches the part close to the joint axis first (which does not occur in the simulation), causing a huge rotation of the handle.
    \item The dynamics properties of the real articulated objects are complicated. For example, the elastic deformation of laptops is not modeled in the simulation.
    \item The kinematic structure of a real articulated object is not ideal. For example, there might be gaps in the drawer rails, which turns the original prismatic joint into a joint with several DOFs.
\end{enumerate}

\subsection{Ablation Study}
\begin{figure}[thpb]
      \centering
      \includegraphics[width=5cm]{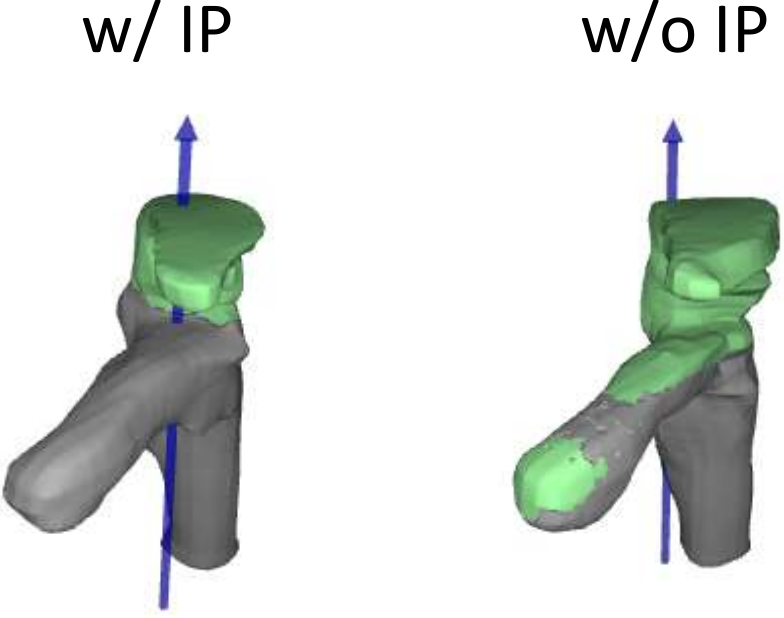}
      \caption{Ablation study on Interactive Perception. The model constructed with Interactive Perception (IP) has more accurate movable part segmentation and joint axis estimation than without IP.}
      \label{fig:ditto}
      \vspace{-5mm}
\end{figure}

\noindent{}\textbf{Interactive Perception.}
To evaluate the necessity of the Interactive Perception module, we train the ablated version, which only uses a static single-frame point cloud. For a fair comparison, we do not change the network architecture and use two same point clouds as the network input. The loss related to the joint state is removed. Fig.~\ref{fig:ditto} shows the comparison result of a real faucet. By actively interacting with the object, we can build a model with more accurate movable part segmentation and joint axis estimation, which is necessary for precise manipulation.

\begin{figure}[!thpb]
      \centering
      \includegraphics[width=8cm]{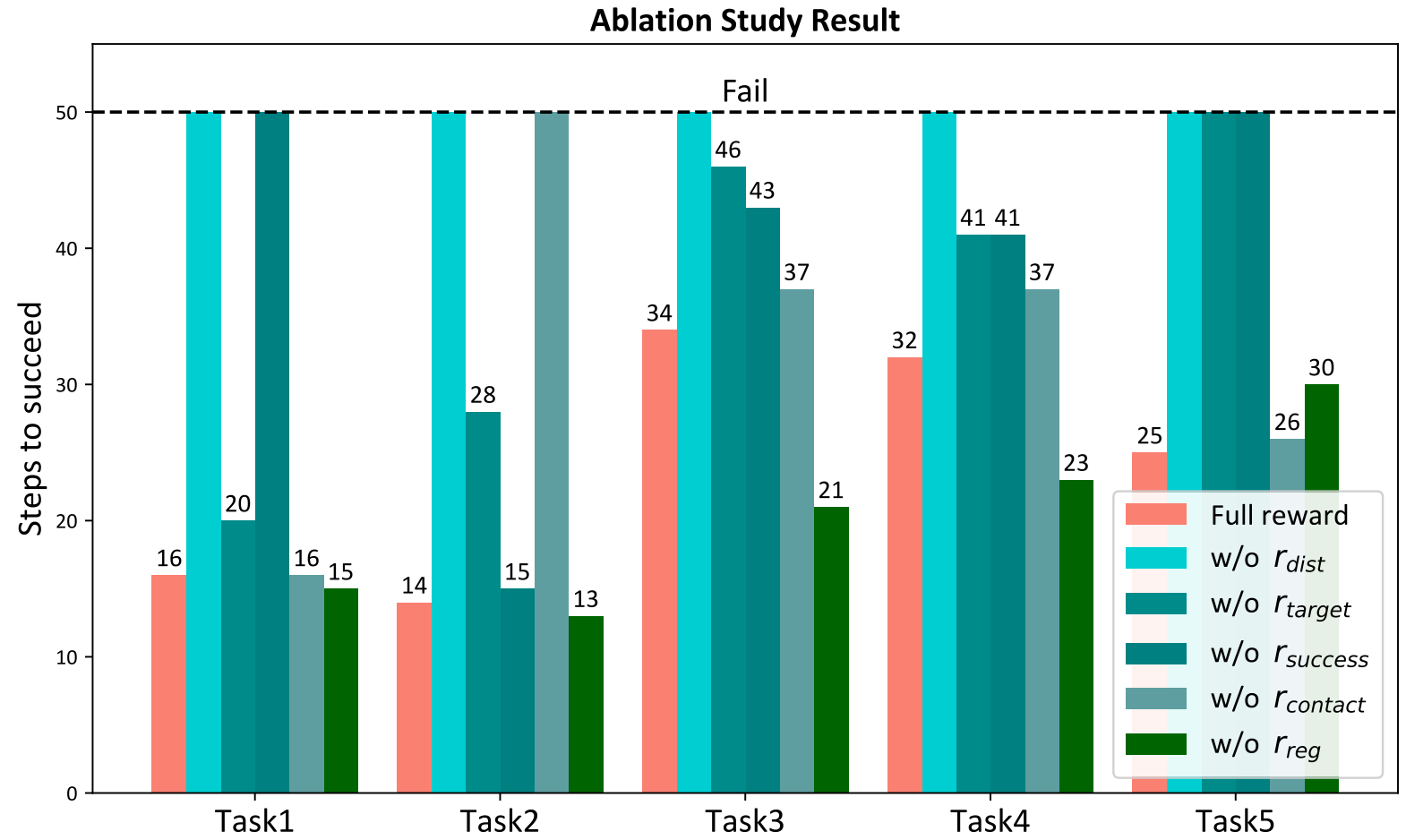}
      \vspace{-3mm}
      \caption{Results of ablation study on reward function. The 5 tasks are opening/closing drawer, opening/closing laptop and turning faucet. The task is considered to be failed if it is not done when time step reaches 50.}
      \label{fig:ablation_study}
      \vspace{-5mm}
\end{figure}

\noindent{}\textbf{Reward function.}
The reward function in the sampling-based model predictive control module is designed to guide the robot to complete the task. To examine the influence of each term of the reward function, we design this ablation study. There are 5 terms in the reward function, so 6 groups of experiments are conducted to reveal each term's influence against the full reward function. The first group runs iCEM with the full reward function as in~\ref{Sampling Based Model Predictive Control}. Each of the other 5 groups drops one term of the full reward function. In each group, 5 tasks are conducted to make the results more general. The task that is not done when the time step reaches 50 is considered to be failed. Fig.~\ref{fig:ablation_study} summarizes the experimental results. 

The experiments using the full reward function are superior in both success rate and steps to succeed, except for the experiments without $r_{reg}$. However, the trajectories searched in w/o $r_{reg}$ are not suitable for real execution, because the robot tends to move to an unusual configuration which could be dangerous when executed in the real world. Without $r_{dist}$, it's impossible for the robot to complete the task because the horizon is too short for the robot to get a positive reward. Without $r_{target}$, $r_{success}$, or $r_{contact}$, the success rates are lowered, and for successful cases, the robot needs more steps to complete the task.

\subsection{Advanced Manipulation Skills}
\begin{figure}[thpb]
    \centering
    \includegraphics[width=8.5cm]{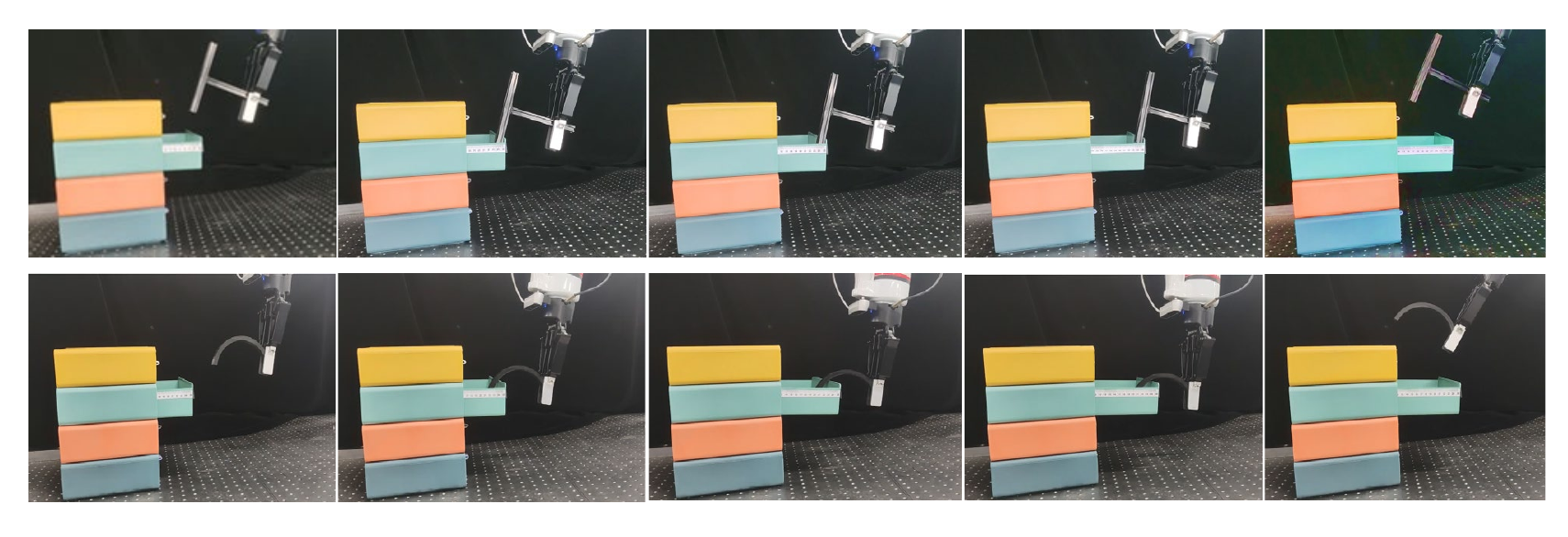}
    \caption{Open drawer with tools. In real scenarios, the object may be beyond the robot's reach, or the gripper cannot fit into the object's size. Our method can be extended to tool-using cases. As shown in these two sequences, the robot uses a T-shaped tool or a semi-ring to open the small drawer.}
    \label{fig:tool_use}
    \vspace{-2mm}
\end{figure}

Our method can be easily extended to support advanced manipulation skills, such as manipulating with tools. When the drawer is located out of the dexterous range of the robot or the gap between the drawer front and the drawer body is small, it cannot be opened only using the gripper. The robot can use tools around it to help complete the task. 
We choose two tools to complete the drawer opening task to verify the tool-using capability of our method (Fig.~\ref{fig:tool_use}). Benefiting from the explicit physics model, it is not complicated to install a tool on the robot to interact with the articulated object. When using MPC to search for trajectories, we assume the tools are mounted on the robot gripper. We simply change the gripper tips to the tool in $r_{dist}$ when computing rewards. We observe that the robot can find the trajectory successfully with most parameters unchanged. For the semi-ring tool, we increase the sample population $N$ to 600.

\section{CONCLUSIONS}

In this work, we present Sim2Real$^\textbf{2}$, a robot learning framework for precise articulated object manipulation. We first build the explicit physics model of the target object through active interaction and then use MPC to search for a long-horizon manipulation trajectory. Quantitative evaluation of real object manipulation results verifies the effectiveness of our proposed framework. For future work, we plan to integrate proprioceptive sensing during real-robot interaction to refine the constructed model for more precise manipulation. Besides, a module that estimates the state of the object in real time will be helpful for reactive and multiple manipulations. Also, we hope to extend the framework to objects with multiple movable parts.

\bibliographystyle{IEEEtran}
\bibliography{citation}

\begin{thebibliography}{10}
\providecommand{\url}[1]{#1}
\csname url@rmstyle\endcsname
\providecommand{\newblock}{\relax}
\providecommand{\bibinfo}[2]{#2}
\providecommand\BIBentrySTDinterwordspacing{\spaceskip=0pt\relax}
\providecommand\BIBentryALTinterwordstretchfactor{4}
\providecommand\BIBentryALTinterwordspacing{\spaceskip=\fontdimen2\font plus
\BIBentryALTinterwordstretchfactor\fontdimen3\font minus
  \fontdimen4\font\relax}
\providecommand\BIBforeignlanguage[2]{{%
\expandafter\ifx\csname l@#1\endcsname\relax
\typeout{** WARNING: IEEEtran.bst: No hyphenation pattern has been}%
\typeout{** loaded for the language `#1'. Using the pattern for}%
\typeout{** the default language instead.}%
\else
\language=\csname l@#1\endcsname
\fi
#2}}

\bibitem{mu2021maniskill}
T.~Mu, Z.~Ling, F.~Xiang, D.~Yang, X.~Li, S.~Tao, Z.~Huang, Z.~Jia, and H.~Su,
  ``Maniskill: Generalizable manipulation skill benchmark with large-scale
  demonstrations,'' \emph{arXiv preprint arXiv:2107.14483}, 2021.

\bibitem{johnson2000thinking}
S.~H. Johnson, ``Thinking ahead: the case for motor imagery in prospective
  judgements of prehension,'' \emph{Cognition}, vol.~74, no.~1, pp. 33--70,
  2000.

\bibitem{von2007action}
C.~Von~Hofsten, ``Action in development,'' \emph{Developmental science},
  vol.~10, no.~1, pp. 54--60, 2007.

\bibitem{hofsten2009action}
C.~V. Hofsten, ``Action, the foundation for cognitive development,''
  \emph{Scandinavian Journal of Psychology}, vol.~50, no.~6, pp. 617--623,
  2009.

\bibitem{hafner2019dream}
D.~Hafner, T.~Lillicrap, J.~Ba, and M.~Norouzi, ``Dream to control: Learning
  behaviors by latent imagination,'' \emph{arXiv preprint arXiv:1912.01603},
  2019.

\bibitem{hafner2020mastering}
D.~Hafner, T.~Lillicrap, M.~Norouzi, and J.~Ba, ``Mastering atari with discrete
  world models,'' \emph{arXiv preprint arXiv:2010.02193}, 2020.

\bibitem{wu2022daydreamer}
P.~Wu, A.~Escontrela, D.~Hafner, K.~Goldberg, and P.~Abbeel, ``Daydreamer:
  World models for physical robot learning,'' \emph{arXiv preprint
  arXiv:2206.14176}, 2022.

\bibitem{polydoros2017survey}
A.~S. Polydoros and L.~Nalpantidis, ``Survey of model-based reinforcement
  learning: Applications on robotics,'' \emph{Journal of Intelligent \& Robotic
  Systems}, vol.~86, no.~2, pp. 153--173, 2017.

\bibitem{wang2019benchmarking}
T.~Wang, X.~Bao, I.~Clavera, J.~Hoang, Y.~Wen, E.~Langlois, S.~Zhang, G.~Zhang,
  P.~Abbeel, and J.~Ba, ``Benchmarking model-based reinforcement learning,''
  \emph{arXiv preprint arXiv:1907.02057}, 2019.

\bibitem{moerland2020model}
T.~M. Moerland, J.~Broekens, and C.~M. Jonker, ``Model-based reinforcement
  learning: A survey,'' \emph{arXiv preprint arXiv:2006.16712}, 2020.

\bibitem{plaat2020model}
A.~Plaat, W.~Kosters, and M.~Preuss, ``Model-based deep reinforcement learning
  for high-dimensional problems, a survey,'' \emph{arXiv preprint
  arXiv:2008.05598}, 2020.

\bibitem{james2020rlbench}
S.~James, Z.~Ma, D.~R. Arrojo, and A.~J. Davison, ``Rlbench: The robot learning
  benchmark \& learning environment,'' \emph{IEEE Robotics and Automation
  Letters}, vol.~5, no.~2, pp. 3019--3026, 2020.

\bibitem{zhu2020robosuite}
Y.~Zhu, J.~Wong, A.~Mandlekar, and R.~Mart{\'\i}n-Mart{\'\i}n, ``robosuite: A
  modular simulation framework and benchmark for robot learning,'' \emph{arXiv
  preprint arXiv:2009.12293}, 2020.

\bibitem{makoviychuk2021isaac}
V.~Makoviychuk, L.~Wawrzyniak, Y.~Guo, M.~Lu, K.~Storey, M.~Macklin,
  D.~Hoeller, N.~Rudin, A.~Allshire, A.~Handa, \emph{et~al.}, ``Isaac gym: High
  performance gpu-based physics simulation for robot learning,'' \emph{arXiv
  preprint arXiv:2108.10470}, 2021.

\bibitem{sadeghi2018sim2real}
F.~Sadeghi, A.~Toshev, E.~Jang, and S.~Levine, ``Sim2real viewpoint invariant
  visual servoing by recurrent control,'' in \emph{Proceedings of the IEEE
  Conference on Computer Vision and Pattern Recognition}, 2018, pp. 4691--4699.

\bibitem{hofer2021sim2real}
S.~H{\"o}fer, K.~Bekris, A.~Handa, J.~C. Gamboa, M.~Mozifian, F.~Golemo,
  C.~Atkeson, D.~Fox, K.~Goldberg, J.~Leonard, \emph{et~al.}, ``Sim2real in
  robotics and automation: Applications and challenges,'' \emph{IEEE
  transactions on automation science and engineering}, vol.~18, no.~2, pp.
  398--400, 2021.

\bibitem{dimitropoulos2022brief}
K.~Dimitropoulos, I.~Hatzilygeroudis, and K.~Chatzilygeroudis, ``A brief survey
  of sim2real methods for robot learning,'' in \emph{International Conference
  on Robotics in Alpe-Adria Danube Region}.\hskip 1em plus 0.5em minus
  0.4em\relax Springer, 2022, pp. 133--140.

\bibitem{gadre2021act}
S.~Y. Gadre, K.~Ehsani, and S.~Song, ``Act the part: Learning interaction
  strategies for articulated object part discovery,'' in \emph{Proceedings of
  the IEEE/CVF International Conference on Computer Vision}, 2021, pp.
  15\,752--15\,761.

\bibitem{lv2022sagci}
J.~Lv, Q.~Yu, L.~Shao, W.~Liu, W.~Xu, and C.~Lu, ``Sagci-system: Towards
  sample-efficient, generalizable, compositional, and incremental robot
  learning,'' in \emph{2022 IEEE International Conference on Robotics and
  Automation (ICRA)}.\hskip 1em plus 0.5em minus 0.4em\relax IEEE, 2022.

\bibitem{xiang2020sapien}
F.~Xiang, Y.~Qin, K.~Mo, Y.~Xia, H.~Zhu, F.~Liu, M.~Liu, H.~Jiang, Y.~Yuan,
  H.~Wang, \emph{et~al.}, ``Sapien: A simulated part-based interactive
  environment,'' in \emph{Proceedings of the IEEE/CVF Conference on Computer
  Vision and Pattern Recognition}, 2020, pp. 11\,097--11\,107.

\bibitem{pinneri2020sample}
C.~Pinneri, S.~Sawant, S.~Blaes, J.~Achterhold, J.~Stueckler, M.~Rolinek, and
  G.~Martius, ``Sample-efficient cross-entropy method for real-time planning,''
  \emph{arXiv preprint arXiv:2008.06389}, 2020.

\bibitem{mo2021where2act}
K.~Mo, L.~J. Guibas, M.~Mukadam, A.~Gupta, and S.~Tulsiani, ``Where2act: From
  pixels to actions for articulated 3d objects,'' in \emph{Proceedings of the
  IEEE/CVF International Conference on Computer Vision}, 2021, pp. 6813--6823.

\bibitem{eisner2022flowbot3d}
B.~Eisner, H.~Zhang, and D.~Held, ``Flowbot3d: Learning 3d articulation flow to
  manipulate articulated objects,'' \emph{arXiv preprint arXiv:2205.04382},
  2022.

\bibitem{xu2021umpnet}
Z.~Xu, Z.~He, and S.~Song, ``Umpnet: Universal manipulation policy network for
  articulated objects,'' \emph{arXiv preprint arXiv:2109.05668}, 2021.

\bibitem{jiang2022ditto}
Z.~Jiang, C.-C. Hsu, and Y.~Zhu, ``Ditto: Building digital twins of articulated
  objects from interaction,'' in \emph{arXiv preprint arXiv:2202.08227}, 2022.

\bibitem{zhang2022close}
X.~Zhang, R.~Chen, F.~Xiang, Y.~Qin, J.~Gu, Z.~Ling, M.~Liu, P.~Zeng, S.~Han,
  Z.~Huang, \emph{et~al.}, ``Close the visual domain gap by physics-grounded
  active stereovision depth sensor simulation,'' \emph{arXiv preprint
  arXiv:2201.11924}, 2022.

\bibitem{mamou2009simple}
K.~Mamou and F.~Ghorbel, ``A simple and efficient approach for 3d mesh
  approximate convex decomposition,'' in \emph{2009 16th IEEE international
  conference on image processing (ICIP)}.\hskip 1em plus 0.5em minus
  0.4em\relax IEEE, 2009, pp. 3501--3504.

\bibitem{wang2019shape2motion}
X.~Wang, B.~Zhou, Y.~Shi, X.~Chen, Q.~Zhao, and K.~Xu, ``Shape2motion: Joint
  analysis of motion parts and attributes from 3d shapes,'' in
  \emph{Proceedings of the IEEE/CVF Conference on Computer Vision and Pattern
  Recognition}, 2019, pp. 8876--8884.

\bibitem{huang2020manifoldplus}
J.~Huang, Y.~Zhou, and L.~Guibas, ``Manifoldplus: A robust and scalable
  watertight manifold surface generation method for triangle soups,''
  \emph{arXiv preprint arXiv:2005.11621}, 2020.

\end{thebibliography}
\end{document}